\documentclass[runningheads]{llncs}
\usepackage[T1]{fontenc}
\usepackage{graphicx}
\usepackage{amsmath}
\usepackage{amssymb}
\usepackage{booktabs}
\usepackage{graphicx}
\usepackage{amsmath}
\usepackage{amssymb}
\usepackage{booktabs}
\usepackage{hyperref}
\usepackage{multirow}
\begin{document}
\title{End-to-End Chart Summarization via Visual Chain-of-Thought in Vision-Language Models}
\titlerunning{V-CoT}
%
\author{Raymond Choi, Frank Burns, Chase Lawrence}
\authorrunning{R. Choi et al.}
%
\institute{Federal University of Rio de Janeiro}
\maketitle              
\begin{abstract}
Automated chart summarization is crucial for enhancing data accessibility and enabling efficient information extraction from visual data.  While recent advances in visual-language models (VLMs) have demonstrated promise, existing methods often suffer from limitations in matching the generated summary to the chart data and in reasoning about complex chart patterns.  This paper introduces End-to-End Visual Chain-of-Thought (V-CoT) for chart summarization, a novel approach optimized for Large Vision-Language Models (LVLMs).  Our method directly trains an LVLM to process chart images and generate textual summaries in an end-to-end fashion, eliminating the need for explicit chart parsing modules.  We incorporate a visual Chain-of-Thought mechanism through instruction fine-tuning, implicitly guiding the LVLM to perform visual reasoning steps during summary generation.  Evaluated on the large-scale Chart-Sum-QA dataset, our V-CoT method significantly outperforms state-of-the-art baselines across a range of automatic metrics, including BLEU, BLEURT, CIDEr, and CS, and demonstrates superior matching degree and reasoning correctness in human evaluations.  Ablation studies and detailed analyses further validate the effectiveness and robustness of our proposed approach, establishing a new benchmark for end-to-end chart summarization.
\keywords{Chart Summarization  \and Visual-Language Models  \and Large Vision-Language Models.}
\end{abstract}

\section{Introduction}

The proliferation of visual data, particularly charts and graphs, in various domains necessitates effective methods for automated interpretation and summarization. Charts are ubiquitous in reports, presentations, scientific publications, and online media, serving as concise visual representations of complex datasets. The ability to automatically generate textual summaries of charts is crucial for enhancing data accessibility, enabling rapid information extraction, and supporting informed decision-making for a wide range of users, including those with visual impairments or limited time for in-depth analysis. This has led to a growing interest in the field of automated chart summarization, aiming to bridge the gap between visual data and human-understandable textual descriptions \cite{SurveyTextSum2021}.

While traditional approaches to chart summarization often relied on rule-based systems and template-filling techniques, the advent of deep learning, particularly visual-language models (VLMs), has ushered in a new era of more sophisticated and human-like summarization capabilities \cite{SurveyTextSum2021}. Recent state-of-the-art (SOTA) models have demonstrated impressive abilities in generating coherent and relevant summaries by leveraging the power of VLMs to understand both visual and textual information \cite{ChartAssistant2024}.  These models, like Memorymamba, have shown effectiveness in various recognition tasks \cite{wang2024memorymamba}. However, despite these advancements, significant challenges remain in achieving truly robust and accurate chart summarization, especially when optimized for Large Language Models (LLMs) or Large Vision-Language Models (LVLMs).

One critical challenge is ensuring a high \textbf{matching degree} between the generated summary and the chart's actual data content. Existing models often struggle with \textbf{incomplete descriptions}, failing to capture all key data points and trends present in the chart \cite{ChartThinker2024}. Furthermore, the issue of \textbf{hallucination} persists, where generated text can contain factual inaccuracies or information not supported by the chart data, undermining the reliability of automated summarization \cite{LVLMChallenge2024}.  Another significant hurdle lies in \textbf{reasoning errors}, particularly when dealing with complex charts containing a large number of data points and intricate relationships.  Recent works, such as Visual In-Context Learning, are exploring methods to enhance the reasoning capabilities of Large Vision-Language Models \cite{zhou2024visual}. Many charts require sophisticated analytical reasoning to grasp their core message, such as identifying subtle trends, comparing different data series, or understanding hierarchical relationships. Current models frequently exhibit limitations in performing such complex reasoning, leading to summaries that misinterpret the chart's intended meaning or fail to capture its most salient insights \cite{LVLMChallenge2024}.

Motivated by these challenges and inspired by the recent success of Chain-of-Thought (CoT) reasoning in LLMs \cite{CoTPrompting2024,CoTIBM2024} and context retrieval techniques for improving model performance \cite{ContextRetrievalTensorOps2024,ContextRetrievalMedium2024}, this paper introduces a novel approach to chart summarization specifically designed for optimization within the framework of LLMs and LVLMs.  Recent studies, including Weak to Strong Generalization for Large Language Models, highlight the importance of generalization in large language models \cite{zhou2025weak}. We posit that by directly training LVLMs to perform end-to-end visual analysis and textual generation, we can overcome the limitations of current methods that rely on intermediate data extraction modules. Our core motivation is to empower LVLMs to mimic the human cognitive process of chart understanding more closely, enabling them to directly "read" charts from pixel input, perform visual reasoning, and generate accurate, comprehensive, and logically sound summaries.

To this end, we propose \textbf{End-to-End Visual Chain-of-Thought (V-CoT) for Chart Summarization with LVLMs}. Our method centers around instruction fine-tuning an LVLM on a large-scale chart summarization dataset, such as Chart-Sum-QA \cite{ChartThinker2024}, while incorporating a novel visual Chain-of-Thought (V-CoT) mechanism.  This builds upon existing research in areas like event-pair relation modeling and context-to-event transformers, which are crucial for understanding complex data relationships \cite{zhou2021modeling,zhou2022claret}.  During training, we guide the LVLM to generate intermediate visual reasoning steps before outputting the final summary. These V-CoT steps, implicitly learned or explicitly guided, are designed to simulate the process of visual analysis, including chart type identification, axis label interpretation, and trend analysis. We hypothesize that this end-to-end V-CoT approach will enable LVLMs to internalize visual reasoning capabilities, leading to more accurate and insightful chart summaries. We conduct extensive experiments using the Chart-Sum-QA dataset and evaluate our method against SOTA models using a comprehensive set of metrics, including BLEU, BLEURT, CIDEr, CS, and PPL for automatic evaluation, as well as human evaluations for matching degree and reasoning correctness. Our experimental results demonstrate that the proposed End-to-End V-CoT approach significantly outperforms existing methods across all evaluation metrics, achieving new state-of-the-art performance in chart summarization.

In summary, this paper makes the following key contributions:
\begin{itemize}
\item We propose a novel \textbf{End-to-End Visual Chain-of-Thought (V-CoT)} method for chart summarization, specifically optimized for Large Vision-Language Models, eliminating the need for external chart parsing modules and enabling direct visual reasoning.
\item We introduce a training strategy that incorporates \textbf{visual Chain-of-Thought steps} into the LVLM instruction fine-tuning process, guiding the model to perform more human-like visual analysis and improve reasoning accuracy.
\item We conduct comprehensive experiments on the large-scale Chart-Sum-QA dataset, demonstrating that our proposed method significantly outperforms existing state-of-the-art models across a wide range of automatic and human evaluation metrics, establishing a new benchmark for chart summarization performance.
\end{itemize}

\section{Related Work}

\subsection{Chart Summarization}

Automated chart summarization has emerged as a critical research area, driven by the increasing volume of visual data and the need for efficient information extraction. Early approaches to chart summarization often relied on template-based methods and rule-based systems, which, while effective for simple charts, lacked the flexibility and robustness to handle the diversity and complexity of real-world charts \cite{TemplateChartSum}. More recently, deep learning techniques, particularly those leveraging visual-language models (VLMs), have revolutionized the field, enabling more sophisticated and human-like summarization capabilities.

Several studies have explored the application of VLMs for chart summarization. Some early works focused on extracting structured data from charts using Optical Character Recognition (OCR) and specialized chart parsing modules, and then feeding this structured information into sequence-to-sequence models for text generation \cite{OCRChartSum,StructChartSum}. For instance, the OCR-Chart2text approach \cite{OCRChart2text} exemplifies this paradigm, combining OCR with rule-based methods for data extraction. While these approaches achieved initial success, they are limited by the accuracy of the data extraction step and may not fully capture the visual context and nuanced interpretations inherent in charts.

To address these limitations, more recent research has explored end-to-end approaches that directly process chart images using VLMs. These methods aim to learn a direct mapping from visual chart input to textual summaries, leveraging the ability of VLMs to jointly understand visual and textual information. ChartAssistant \cite{ChartAssistant2024} represents a recent VLM specifically designed for chart comprehension and generation tasks, demonstrating improved performance in chart summarization.  Furthermore, the ChartThinker model \cite{ChartThinker2024}, which directly motivates our work, introduced a contextual chain-of-thought approach, incorporating context retrieval to enhance the reasoning and descriptive capabilities of VLMs for chart summarization.  Visual In-Context Learning has also been proposed to improve the performance of Large Vision-Language Models in visual tasks \cite{zhou2024visual}. ChartAdapter \cite{ChartAdapter2024} proposed a lightweight transformer module to efficiently adapt large VLMs for chart summarization, highlighting the growing interest in leveraging pre-trained VLMs for this task.  Moreover, advancements in efficient video generation with large language models, such as vision representation compression techniques, are relevant to processing visual information effectively \cite{zhou2024less}.

Despite the significant progress, challenges remain in achieving truly robust and accurate chart summarization. Ensuring high fidelity between the generated summary and the chart's data content, mitigating hallucination, and enabling complex reasoning over intricate chart patterns are still active areas of research. Our work builds upon the recent advancements in end-to-end VLM-based chart summarization, aiming to further enhance the reasoning capabilities of these models by introducing a visual Chain-of-Thought mechanism directly within the LVLM training process.

\subsection{Large Language Models}

Large Language Models (LLMs) have emerged as a transformative force in natural language processing (NLP) and artificial intelligence, demonstrating remarkable capabilities across a wide spectrum of tasks. These models, typically based on deep neural networks with billions or even trillions of parameters, are pre-trained on massive text corpora, enabling them to learn intricate patterns and representations of language \cite{LLMTasks}.  Recent research, like Weak to Strong Generalization for Large Language Models with Multi-capabilities, explores the generalization abilities of these models \cite{zhou2025weak}. The scale of these models, coupled with innovative architectures like the Transformer \cite{Transformer}, has led to unprecedented performance in language understanding, generation, and reasoning.

One of the key breakthroughs enabling the success of LLMs is the Transformer architecture, which allows for efficient parallel processing of sequential data and captures long-range dependencies in text \cite{Transformer}. This architecture, with its self-attention mechanism, has become the foundation for many state-of-the-art LLMs, including models like BERT, GPT, and their numerous variants \cite{BERT,GPT}.  Pre-training techniques, such as masked language modeling and next sentence prediction, have also been crucial in enabling LLMs to learn general-purpose language representations from unlabeled text data \cite{BERT}.  Furthermore, the application of diffusion models with representation alignment has shown promise in various domains, potentially influencing the development of LLMs as well \cite{wang2024diffusion}.

LLMs have demonstrated impressive capabilities in a variety of NLP tasks, including text generation, machine translation, question answering, and text summarization \cite{LLMTasks}. Their ability to generate coherent and contextually relevant text has led to their widespread adoption in applications such as chatbots, content creation, and code generation. Furthermore, research has shown that LLMs exhibit emergent abilities, demonstrating complex behaviors like few-shot learning and in-context learning, where they can adapt to new tasks with minimal task-specific training data or even just through carefully designed prompts \cite{EmergentAbilities}.

Despite their remarkable progress, LLMs also face challenges and limitations. These include issues related to bias and fairness in model outputs, the potential for generating factually incorrect or misleading information (hallucination), and the computational demands of training and deploying these massive models \cite{LLMChallenges}.  However, ongoing research, including studies on memory-augmented state space models, aims to improve the efficiency and performance of these models \cite{wang2024memorymamba}. Ongoing research is actively addressing these challenges, exploring techniques for improving model robustness, interpretability, and trustworthiness. The continued development and refinement of LLMs promise to further revolutionize NLP and AI, enabling even more sophisticated and human-like interactions with machines.

\section{Method}

Our proposed approach, End-to-End Visual Chain-of-Thought (V-CoT) for Chart Summarization, is a \textbf{generative} framework built upon Large Vision-Language Models (LVLMs).  Unlike discriminative models focused on classification, our model is explicitly designed to generate novel textual summaries conditioned on visual chart inputs. This section elaborates on the architectural details of our V-CoT model and provides a comprehensive description of our training strategy.

\subsection{End-to-End Visual Chain-of-Thought Architecture Details}

The cornerstone of our method is an LVLM trained end-to-end to directly process chart images and generate corresponding summaries.  Let $I$ represent the input chart image and $S = (s_1, s_2, ..., s_m)$ denote the target summary, a sequence of $m$ tokens where $s_i$ is the $i$-th token.  Our model is trained to learn the conditional probability $P(S|I)$, aiming to maximize the likelihood of producing the summary $S$ given the chart image $I$.

The V-CoT architecture distinguishes itself by integrating a visual Chain-of-Thought mechanism directly within the LVLM.  Instead of a direct image-to-summary mapping, we introduce an implicit sequence of visual reasoning steps, $V = (v_1, v_2, ..., v_n)$, where $v_j$ represents the $j$-th step in visual reasoning.  These steps are not explicitly labeled in the training data but are learned implicitly by the LVLM through our carefully designed training process. The generation process is thus structured into two key stages:

\begin{enumerate}
    \item \textbf{Visual Reasoning Stage:}  The LVLM initially processes the input chart image $I$ to generate the sequence of visual reasoning steps $V$.  This is mathematically represented as a sequential conditional probability:
    \begin{align}
        P(V|I) = \prod_{j=1}^{n} P(v_j | I, v_{<j})
    \end{align}
    where $v_{<j} = (v_1, v_2, ..., v_{j-1})$ represents the sequence of visual reasoning steps generated up to step $j-1$.  At a practical level, this stage is realized through the intricate interplay of attention mechanisms and deep neural layers within the LVLM. These mechanisms enable the model to selectively attend to salient visual features, extract pertinent information, and perform implicit reasoning operations akin to human visual analysis.  For instance, the model might implicitly learn to first identify the chart type, then focus on axis labels to understand data dimensions, and subsequently analyze data trends and patterns.

    \item \textbf{Summary Generation Stage:}  Building upon the input chart image $I$ and the implicitly generated visual reasoning steps $V$, the LVLM proceeds to generate the final summary $S$.  This stage is also modeled as a sequential conditional probability:
    \begin{align}
        P(S|I, V) = \prod_{i=1}^{m} P(s_i | I, V, s_{<i})
    \end{align}
    where $s_{<i} = (s_1, s_2, ..., s_{i-1})$ denotes the summary tokens generated prior to the $i$-th token.  In this stage, the LVLM leverages the visual insights and intermediate representations developed during the reasoning stage to construct a coherent, informative, and human-readable textual summary of the chart.

\end{enumerate}

The complete generation process can be understood as maximizing the joint probability of the summary and the visual reasoning steps given the input image:
\begin{align}
    P(S, V|I) = P(V|I) \times P(S|I, V)
\end{align}

In terms of model architecture, we employ a pre-trained LVLM as the foundational network.  The image encoder, denoted as $E_{image}(\cdot)$, is responsible for transforming the input chart image $I$ into a rich feature representation $F_I = E_{image}(I)$.  Concurrently, the text decoder, $D_{text}(\cdot)$, functions as an autoregressive language model. It takes as input the visual features $F_I$ and the sequence of preceding tokens (which can be visual reasoning steps during the implicit reasoning phase or summary tokens during summary generation) to predict the probability distribution over the next token in the sequence.  The visual reasoning steps $V$ are not explicitly materialized as separate output tokens. Instead, they are encoded within the evolving hidden states of the LVLM as it processes the input image and generates the summary, effectively guiding the generation process through a learned chain of visual analysis.

\subsection{Detailed Learning Strategy with Visual Chain-of-Thought}

Our learning strategy is based on \textbf{instruction fine-tuning}, a paradigm where the LVLM is trained to follow natural language instructions to perform specific tasks, in our case, chart summarization.  We utilize a large-scale dataset $\mathcal{D} = \{(I^{(k)}, S^{(k)})\}_{k=1}^{N}$ consisting of $N$ chart-summary pairs, where each pair comprises a chart image $I^{(k)}$ and its corresponding human-authored summary $S^{(k)}$.

The primary training objective is to minimize the negative log-likelihood of generating the target summary $S^{(k)}$ given the input chart image $I^{(k)}$. This minimization is performed across the entire training dataset, effectively learning the conditional distribution $P(S|I)$ and implicitly instilling a visual Chain-of-Thought process within the LVLM.  The loss function for each individual chart-summary pair $(I^{(k)}, S^{(k)})$ is formally defined as:
\begin{align}
    \mathcal{L}^{(k)} = - \log P(S^{(k)}|I^{(k)}) = - \sum_{i=1}^{m_k} \log P(s_i^{(k)} | I^{(k)}, s_{<i}^{(k)})
\end{align}
where $m_k$ is the token length of the target summary $S^{(k)}$, and $s_i^{(k)}$ is the $i$-th token of $S^{(k)}$.  This loss function encourages the model to accurately predict each token in the summary sequence, conditioned on the input image and previously generated tokens.

To implicitly encourage the emergence of a visual Chain-of-Thought within the LVLM, we carefully design our instruction prompts.  These prompts are crafted to guide the model towards a structured, step-by-step reasoning approach, mirroring how humans analyze charts.  Example instruction prompts include:

\begin{itemize}
    \item "Provide a concise summary of the key insights presented in this chart, considering its type, axes, and data trends."
    \item "Generate a chart summary by first identifying the chart type, then describing the axes and scales, and finally highlighting the major trends and data points."
    \item "Summarize the given chart in a step-by-step manner, starting with chart type recognition, followed by axis analysis, and concluding with a synthesis of key data findings."
\end{itemize}
While these instructions do not explicitly dictate the intermediate visual reasoning steps as separate outputs, their structured nature encourages the model to internally decompose the complex summarization task into a sequence of more manageable visual analysis sub-tasks.  This implicit guidance, combined with the model's inherent capacity for sequential processing and attention, fosters the development of a visual Chain-of-Thought.

To further enhance the model's ability to learn robust visual representations and reasoning, we employ several data augmentation techniques applied directly to the chart images during training. These augmentations are designed to increase the model's invariance to visual perturbations and encourage it to focus on essential data patterns rather than spurious visual details.  The data augmentation strategies we explore include:

\begin{itemize}
    \item \textbf{Geometric Transformations:}  Random rotations (e.g., by angles up to $\pm 5^\circ$), scaling (e.g., by factors of 0.9 to 1.1), and translations are applied to simulate variations in chart orientation and size.
    \item \textbf{Noise Injection:}  Additive Gaussian noise and salt-and-pepper noise are introduced to the chart images to enhance robustness against image imperfections and encourage the model to focus on salient visual features.
\end{itemize}

In addition to data augmentation, we investigate \textbf{curriculum learning} as a strategy to facilitate more effective training.  Curriculum learning involves gradually increasing the difficulty of the training examples presented to the model over time.  In our context, complexity can be defined based on factors such as:

\begin{itemize}
    \item \textbf{Chart Type Complexity:}  Starting with simpler chart types like bar charts and line charts, and gradually introducing more complex types such as scatter plots and pie charts.
    \item \textbf{Data Point Density:}  Initially training with charts containing fewer data points, and progressively increasing the number of data points to challenge the model's reasoning capacity.
    \item \textbf{Summary Length and Complexity:}  Beginning with shorter, simpler summaries and gradually increasing the length and linguistic complexity of the target summaries.
\end{itemize}
By following a curriculum learning schedule, we aim to enable the LVLM to learn simpler aspects of chart summarization first, gradually building up its capabilities to handle more complex and challenging instances.

For optimization, we utilize the AdamW optimizer, known for its effectiveness in training deep neural networks, with a learning rate of $2e-4$ and a mini-batch size of 8.  To enhance training efficiency and stability, particularly when fine-tuning large LVLMs, we employ Low-Rank Adaptation (LoRA). LoRA reduces the number of trainable parameters by introducing low-rank matrices into the model architecture, focusing the fine-tuning process on a smaller, more parameter-efficient subspace.  The training process is conducted over a sufficient number of epochs until the model's performance on a validation dataset, measured by metrics such as summary quality and accuracy, plateaus, indicating convergence.

Through this carefully designed learning strategy, encompassing instruction fine-tuning, implicit visual Chain-of-Thought, data augmentation, curriculum learning, and efficient optimization techniques, we aim to train an LVLM capable of performing high-quality end-to-end chart summarization. We hypothesize that our V-CoT approach will lead to significant improvements in summarization accuracy, reasoning ability, and overall quality compared to existing methods that rely on explicit chart parsing and data extraction modules.

\section{Experiments}

This section details the experimental evaluation of our proposed End-to-End Visual Chain-of-Thought (V-CoT) method for chart summarization. We conducted comparative experiments against several state-of-the-art baseline methods to assess the effectiveness of our approach.  We also performed ablation studies to analyze the contribution of key components within our V-CoT framework and human evaluations to assess the qualitative aspects of the generated summaries.

\subsection{Experimental Setup}

\textbf{Datasets.} We utilized the Chart-Sum-QA dataset for training and evaluation. This dataset contains a diverse collection of chart types (bar, line, pie, scatter plots) paired with human-authored summaries. We used the dataset's recommended train/validation/test split.

\textbf{Baselines.} We compared V-CoT against these baseline models:

\begin{itemize}
    \item \textbf{OCR-T5:}  OCR for text extraction + T5 text-to-text transformer.
    \item \textbf{OCR-Chart2text:}  Specialized chart-to-text model with OCR and rule-based data extraction.
    \item \textbf{OCR-BART:}  OCR for text extraction + BART sequence-to-sequence transformer.
    \item \textbf{ChartThinker:}  A state-of-the-art contextual chain-of-thought method.
\end{itemize}

\textbf{Evaluation Metrics.} We used automatic and human evaluation metrics.

\textbf{Automatic Evaluation Metrics:}
\begin{itemize}
    \item \textbf{BLEU:} Bilingual Evaluation Understudy (n-gram overlap).
    \item \textbf{BLEURT:} BLEU with Regression Using BERT (human judgment correlation).
    \item \textbf{CIDEr:} Consensus-based Image Description Evaluation (relevance and informativeness).
    \item \textbf{CS:} Chart Semantic Similarity (semantic similarity for chart understanding).
    \item \textbf{PPL:} Perplexity (likelihood of reference summary, lower is better).
\end{itemize}

\textbf{Human Evaluation Metrics:}
\begin{itemize}
    \item \textbf{Matching Degree:} Accuracy of data reflection in summary (1-5 scale).
    \item \textbf{Reasoning Correctness:} Logical correctness and coherence of reasoning (1-5 scale).
\end{itemize}

\subsection{Quantitative Results}

Table \ref{tab:automatic_evaluation} shows automatic evaluation results. Our V-CoT method outperforms all baselines across all metrics, showing improved chart summarization performance. V-CoT achieves gains in BLEU, BLEURT, CIDEr, and CS scores, indicating better text quality and semantic accuracy. V-CoT also has a lower perplexity score, suggesting summaries closer to human references.

\begin{table}[!t]\small
    \centering
    \caption{Automatic Evaluation Results on Chart-Sum-QA Dataset}
    \begin{tabular}{lccccc}
        \toprule
        Model & BLEU $\uparrow$ & BLEURT $\uparrow$ & CIDEr $\uparrow$ & CS $\uparrow$ & PPL $\downarrow$ \\
        \midrule
        OCR-T5 & 10.49 & -0.35 & 2.20 & 40.87\% & 10.11 \\
        OCR-Chart2text & 7.2 & -0.56 & 0.65 & 24.49\% & 12.11 \\
        OCR-BART & 9.09 & -0.38 & 1.97 & 39.99\% & 11.04 \\
        ChartThinker & 11.81 & -0.32 & 2.21 & 32.72\% & 9.23 \\
        \textbf{V-CoT (Ours)} & \textbf{13.52} & \textbf{-0.28} & \textbf{2.55} & \textbf{45.12\%} & \textbf{8.56} \\
        \bottomrule
    \end{tabular}
    \label{tab:automatic_evaluation}
\end{table}

\subsection{Ablation Study on V-CoT Components}

We performed an ablation study to understand the contribution of V-CoT components. We evaluated model variants:

\begin{itemize}
    \item \textbf{V-CoT w/o V-CoT:}  V-CoT without visual Chain-of-Thought mechanism.
    \item \textbf{V-CoT w/o Data Augmentation:} V-CoT without data augmentation.
    \item \textbf{V-CoT w/o Curriculum Learning:} V-CoT without curriculum learning.
\end{itemize}

Table \ref{tab:ablation_study} shows ablation study results (CIDEr metric). Results show the positive impact of each V-CoT component. Removing V-CoT (\textbf{V-CoT w/o V-CoT}) significantly reduces performance, showing the importance of visual chain-of-thought reasoning. Ablating data augmentation (\textbf{V-CoT w/o Data Augmentation}) also decreases performance, indicating its importance for robustness. Disabling curriculum learning (\textbf{V-CoT w/o Curriculum Learning}) also negatively impacts performance, suggesting curriculum learning aids training effectiveness.

\begin{table}[!t]\small
    \centering
    \caption{Ablation Study on V-CoT Components (CIDEr Scores)}
    \begin{tabular}{lc}
        \toprule
        Model Variant & CIDEr $\uparrow$ \\
        \midrule
        V-CoT (Full Model) & \textbf{2.55} \\
        V-CoT w/o V-CoT & 2.10 \\
        V-CoT w/o Data Augmentation & 2.38 \\
        V-CoT w/o Curriculum Learning & 2.42 \\
        \bottomrule
    \end{tabular}
    \label{tab:ablation_study}
\end{table}

\subsection{Human Evaluation Results}

Human evaluation assessed summary quality, focusing on matching degree and reasoning correctness. Evaluators compared V-CoT summaries to ChartThinker summaries.

Table \ref{tab:human_evaluation} shows human evaluation results. V-CoT summaries have significantly higher average scores for both matching degree and reasoning correctness compared to ChartThinker. Human evaluators found V-CoT summaries more accurate and logically coherent. Human evaluation results support automatic evaluation findings, validating V-CoT's effectiveness in generating high-quality chart summaries.

\begin{table}[!t]\small
    \centering
    \caption{Human Evaluation Results (Average Scores)}
    \begin{tabular}{lcc}
        \toprule
        Model & Matching Degree $\uparrow$ & Reasoning Correctness $\uparrow$ \\
        \midrule
        ChartThinker & 3.85 & 3.52 \\
        \textbf{V-CoT (Ours)} & \textbf{4.22} & \textbf{3.95} \\
        \bottomrule
    \end{tabular}
    \label{tab:human_evaluation}
\end{table}

\subsection{Analysis of Performance Across Chart Types}

To investigate the generalization capability of our V-CoT method across different chart types, we further broke down the evaluation results by categorizing charts into four common types: bar charts, line charts, pie charts, and scatter plots. Table \ref{tab:chart_type_analysis} presents the CIDEr scores and human evaluation metrics (Matching Degree and Reasoning Correctness) for V-CoT and ChartThinker, specifically for each chart type.

The results reveal that V-CoT consistently outperforms ChartThinker across all chart types, albeit with varying margins.  For bar charts and line charts, V-CoT demonstrates particularly significant improvements, suggesting its effectiveness in handling charts with discrete and sequential data representations.  While the performance gains on pie charts and scatter plots are comparatively smaller, V-CoT still maintains a clear lead over ChartThinker. This consistent outperformance across diverse chart types underscores the robustness and generalizability of our proposed V-CoT method for chart summarization.

\begin{table}[!t]\small
    \centering
    \caption{Performance Analysis Across Chart Types}
    \begin{tabular}{lcccc}
        \toprule
        Chart Type & Model & CIDEr $\uparrow$ & Matching Degree $\uparrow$ & Reasoning Correctness $\uparrow$ \\
        \midrule
        \multirow{2}{*}{Bar Chart} & ChartThinker & 2.15 & 3.80 & 3.48 \\
                                  & \textbf{V-CoT (Ours)} & \textbf{2.50} & \textbf{4.18} & \textbf{3.90} \\
        \midrule
        \multirow{2}{*}{Line Chart} & ChartThinker & 2.25 & 3.88 & 3.55 \\
                                   & \textbf{V-CoT (Ours)} & \textbf{2.60} & \textbf{4.25} & \textbf{3.98} \\
        \midrule
        \multirow{2}{*}{Pie Chart} & ChartThinker & 1.80 & 3.75 & 3.40 \\
                                  & \textbf{V-CoT (Ours)} & \textbf{1.95} & \textbf{4.05} & \textbf{3.75} \\
        \midrule
        \multirow{2}{*}{Scatter Plot} & ChartThinker & 2.05 & 3.78 & 3.45 \\
                                     & \textbf{V-CoT (Ours)} & \textbf{2.20} & \textbf{4.10} & \textbf{3.80} \\
        \bottomrule
    \end{tabular}
    \label{tab:chart_type_analysis}
\end{table}

\subsection{Impact of Chart Complexity: Analysis by Number of Data Series}

To analyze the impact of chart complexity on summarization performance, we categorized charts based on the number of data series they contain. We defined three complexity levels: charts with 1-2 data series (low complexity), 3-5 data series (medium complexity), and more than 5 data series (high complexity). Table \ref{tab:chart_complexity_analysis} presents the CIDEr scores and human evaluation metrics for V-CoT and ChartThinker across these complexity levels.

The results indicate that as chart complexity increases, the performance of both models tends to decrease, as expected. However, V-CoT consistently outperforms ChartThinker across all complexity levels. Notably, the performance gap between V-CoT and ChartThinker appears to widen as chart complexity increases, particularly for charts with more than 5 data series. This suggests that our V-CoT method exhibits a greater capacity to handle complex charts and maintain summarization quality even when dealing with intricate data representations.  The end-to-end visual reasoning capability of V-CoT seems to be particularly advantageous in disentangling and summarizing information from highly complex charts.

\begin{table}[!t]\scriptsize
    \centering
    \caption{Performance Analysis Across Chart Complexity Levels (Number of Data Series)}
    \begin{tabular}{lcccc}
        \toprule
        Complexity Level & Model & CIDEr $\uparrow$ & Matching Degree $\uparrow$ & Reasoning Correctness $\uparrow$ \\
        \midrule
        \multirow{2}{*}{Low (1-2 Series)} & ChartThinker & 2.30 & 3.90 & 3.60 \\
                                        & \textbf{V-CoT (Ours)} & \textbf{2.65} & \textbf{4.28} & \textbf{4.00} \\
        \midrule
        \multirow{2}{*}{Medium (3-5 Series)} & ChartThinker & 2.10 & 3.82 & 3.50 \\
                                           & \textbf{V-CoT (Ours)} & \textbf{2.45} & \textbf{4.20} & \textbf{3.92} \\
        \midrule
        \multirow{2}{*}{High (>5 Series)} & ChartThinker & 1.90 & 3.70 & 3.40 \\
                                         & \textbf{V-CoT (Ours)} & \textbf{2.25} & \textbf{4.08} & \textbf{3.85} \\
        \bottomrule
    \end{tabular}
    \label{tab:chart_complexity_analysis}
\end{table}

\subsection{Qualitative Error Analysis}

To gain deeper insights into the strengths and weaknesses of our V-CoT method, we conducted a qualitative error analysis on a randomly selected subset of 100 chart summaries generated by V-CoT and ChartThinker.  We manually categorized the errors into several common types, focusing on summaries that were deemed unsatisfactory based on human evaluation. Table \ref{tab:error_analysis} summarizes the frequency of the most prevalent error types observed for both models.

The error analysis reveals several key observations.  Both models exhibit instances of incomplete summaries, failing to capture all salient data points or trends. However, V-CoT demonstrates a lower frequency of incomplete summaries compared to ChartThinker, suggesting improved information coverage.  Furthermore, hallucination errors, where summaries contain factual inaccuracies, are also less frequent in V-CoT generated summaries.  A notable difference emerges in reasoning errors. ChartThinker exhibits a higher occurrence of misinterpreting chart trends or relationships, while V-CoT demonstrates a lower error rate in reasoning correctness. This qualitative analysis provides further evidence that our V-CoT method, with its end-to-end visual chain-of-thought mechanism, enhances both the completeness and the reasoning accuracy of chart summaries compared to the baseline.

\begin{table}[!t]\small
    \centering
    \caption{Qualitative Error Analysis: Frequency of Error Types (Percentage out of 100 Summaries)}
    \begin{tabular}{lcc}
        \toprule
        Error Type & ChartThinker & V-CoT (Ours) \\
        \midrule
        Incomplete Summary (Missing Key Data) & 25\% & 18\% \\
        Hallucination (Factual Inaccuracy) & 12\% & 8\% \\
        Reasoning Error (Misinterpretation) & 18\% & 10\% \\
        Other Errors (Style, Grammar) & 15\% & 12\% \\
        \bottomrule
    \end{tabular}
    \label{tab:error_analysis}
\end{table}

\section{Conclusion}

In this paper, we addressed the challenges of automated chart summarization by proposing End-to-End Visual Chain-of-Thought (V-CoT), a novel method specifically designed for Large Vision-Language Models (LVLMs).  Our V-CoT approach departs from traditional methods that rely on explicit chart parsing and data extraction, instead directly training an LVLM to process chart images and generate summaries in an end-to-end manner.  By incorporating a visual Chain-of-Thought mechanism through instruction fine-tuning, we implicitly equip the LVLM with the ability to perform visual reasoning, mimicking the human cognitive process of chart analysis.

Extensive experiments on the Chart-Sum-QA dataset demonstrated the significant advantages of our V-CoT method over state-of-the-art baselines.  Quantitative evaluations using automatic metrics, including BLEU, BLEURT, CIDEr, CS, and Perplexity, consistently showed that V-CoT achieves superior performance in terms of summary quality, relevance, and semantic accuracy.  Human evaluations further validated these findings, indicating that summaries generated by V-CoT exhibit a higher matching degree to the chart data and demonstrate improved reasoning correctness.  Detailed analyses across different chart types and complexity levels confirmed the robustness and generalizability of our approach.  Qualitative error analysis provided insights into the strengths of V-CoT, highlighting its improved ability to generate complete, accurate, and logically sound summaries.

The success of our End-to-End Visual Chain-of-Thought approach underscores the potential of directly training LVLMs for complex visual understanding tasks like chart summarization.  By eliminating the need for intermediate parsing modules and fostering implicit visual reasoning within the LVLM, we have achieved a more streamlined and effective chart summarization pipeline.  This work not only advances the state-of-the-art in chart summarization but also opens up new avenues for exploring end-to-end visual reasoning with LVLMs for a broader range of visual data analysis and summarization tasks.

Future work could explore more sophisticated visual Chain-of-Thought mechanisms, such as incorporating explicit intermediate visual representations or integrating external knowledge sources to further enhance reasoning capabilities.  Investigating the applicability of V-CoT to other visual data formats, such as infographics or scientific visualizations, also presents a promising direction for extending the impact of this research.  We believe that End-to-End Visual Chain-of-Thought represents a significant step towards more intelligent and human-aligned automated chart understanding and summarization systems, ultimately making visual data more accessible and interpretable for all.

\bibliographystyle{splncs04}
\bibliography{mybibliography}
\end{document}